\documentclass[conference]{IEEEtran}
\usepackage{times}
\usepackage{amsmath}
\usepackage{graphicx}
\usepackage[numbers]{natbib}
\usepackage{multicol}
\usepackage[bookmarks=true]{hyperref}
\usepackage{amsmath}
\usepackage{amsfonts}

\DeclareMathOperator*{\argmin}{argmin} 
\usepackage{float}
\usepackage{booktabs}
\usepackage{graphicx}
\usepackage[table,xcdraw]{xcolor}
\usepackage{cuted} 
\usepackage{caption}
\usepackage[ruled,vlined,linesnumbered]{algorithm2e}
\usepackage[noend]{algorithmic}
\usepackage{pifont}

\DeclareRobustCommand{\IEEEauthorrefmark}[1]{\smash{\textsuperscript{\footnotesize #1}}}

\begin{document}

\title{IMLE Policy:\\
\huge{Fast and Sample Efficient Visuomotor Policy Learning via \\Implicit Maximum Likelihood Estimation}}




%

\author{\authorblockN{Krishan Rana$^*$\authorrefmark{1},
Robert Lee$^*$,
David Pershouse\authorrefmark{1} and
Niko S\"{u}nderhauf\authorrefmark{1} }
\authorblockA{$^*$Equal Contribution}
\authorblockA{\authorrefmark{1}QUT Centre for Robotics}
    Email: ranak@qut.edu.au}

\maketitle
\begin{strip}
    \centering
\includegraphics[width=1.0\linewidth]{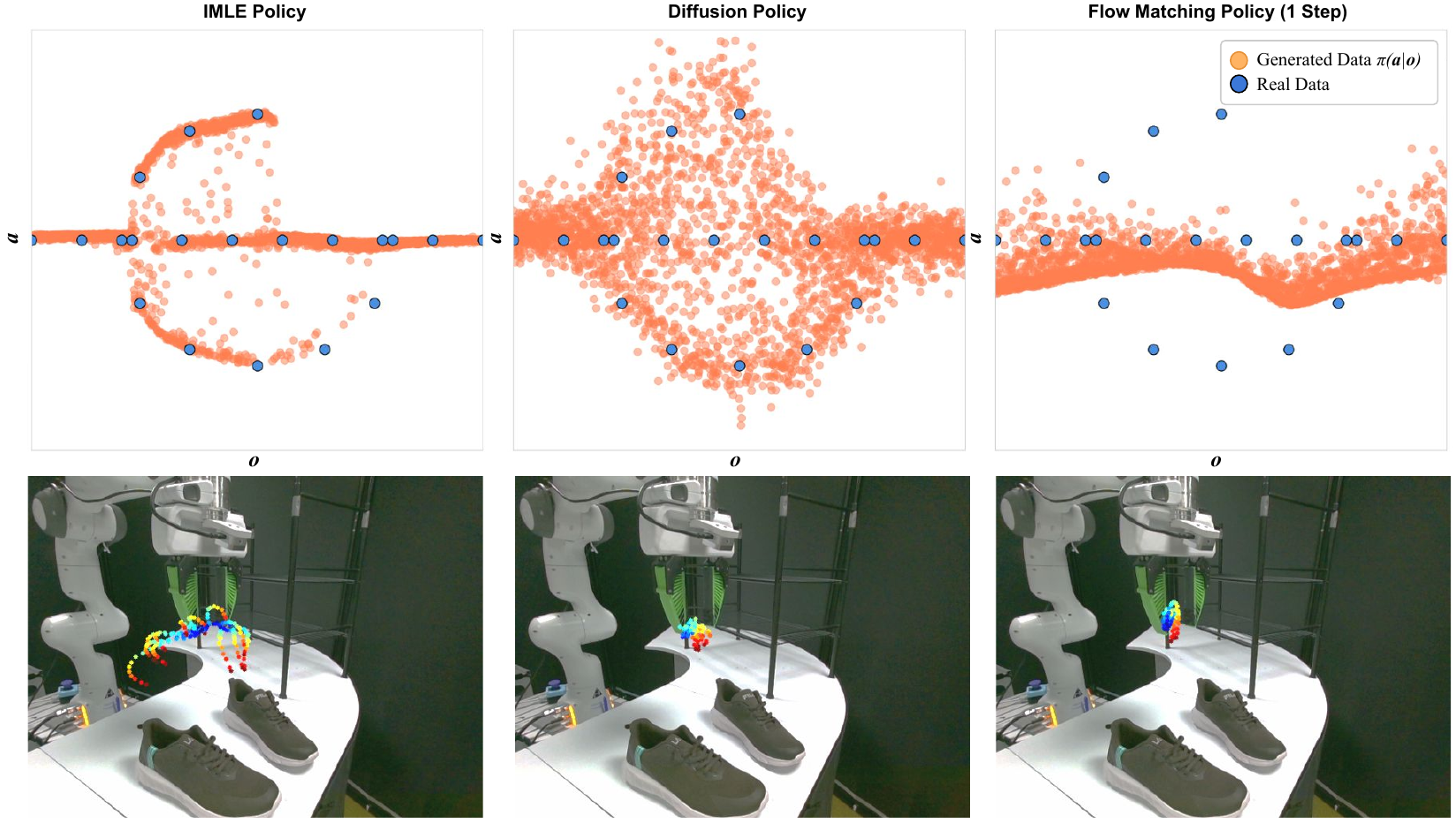}
    \captionof{figure}{
        \textbf{Capturing Multi-Modality from Limited Demonstrations.} We present IMLE Policy, a novel behaviour cloning approach that is capable of learning complex multi-modal action distributions from less data than prior approaches, while only requiring a single network forward pass. \textbf{Top:} When trained on 20 samples for 4000 epochs, IMLE Policy learns to accurately model the underlying multi-modal function, while Diffusion Policy with 100 de-noising steps struggles. Flow Matching with a single step collapses modes. \textbf{Bottom:} When trained on only 17 demonstrations for a shoe racking task, IMLE Policy maintains multi-modality, predicting a variety of trajectories from the starting position where either shoe could be selected first while the other methods predict poor trajectories.
        }
    \label{fig:main}
\end{strip}

\begin{abstract}
Recent advances in imitation learning, particularly using generative modelling techniques like diffusion, have enabled policies to capture complex multi-modal action distributions. However, these methods often require large datasets and multiple inference steps for action generation, posing challenges in robotics where the cost for data collection is high and computation resources are limited. To address this, we introduce IMLE Policy, a novel behaviour cloning approach based on Implicit Maximum Likelihood Estimation (IMLE). IMLE Policy excels in low-data regimes, effectively learning from minimal demonstrations and requiring 38\% less data on average to match the performance of baseline methods in learning complex multi-modal behaviours. Its simple generator-based architecture enables single-step action generation, improving inference speed by 97.3\% compared to Diffusion Policy, while outperforming single-step Flow Matching. We validate our approach across diverse manipulation tasks in simulated and real-world environments, showcasing its ability to capture complex behaviours under data constraints. Videos and code are provided on our project page: \href{https://imle-policy.github.io/}{imle-policy.github.io/}.

\end{abstract}

\IEEEpeerreviewmaketitle

\section{Introduction}

Learning policies from demonstrations is becoming a key approach for enabling robots to perform complex tasks. At its core, this problem involves mapping observations to actions in a way that captures the nuanced and often multi-modal nature of human behaviour. However, leveraging behaviour cloning for real-world robotics applications presents unique challenges including the ability to learn these highly multi-modal action distributions from limited demonstration data which can be expensive to collect, and the need for computational efficiency to enable real-time operation. Recent advancements in generative modelling, such as diffusion models \cite{sohldickstein2015nonequilibrium, chi2023diffusion} and flow matching \cite{lipmanflow, black2024pi_0}, have demonstrated promise in capturing complex multi-modal action distributions for behaviour cloning. However, these approaches are computationally expensive during inference, as they require iterative sampling processes to generate actions. Additionally, they often demand large datasets to train performant models effectively. Addressing these challenges is crucial for enhancing the practicality and adoption of behaviour cloning in real-world tasks.


Generative Adversarial Networks (GANs) are an alternative approach to generative modelling with efficient single-step inference, but fell out of favour due to their issues with training stability and mode collapse, where parts of the data distribution are ignored, limiting their ability to effectively model complex, multi-modal distributions. To address these shortcomings, Implicit Maximum Likelihood Estimation (IMLE) \cite{imle} was introduced as an alternative training objective for single-step generative models, based on a simple objective that ensures every datapoint is well-represented by at least one generated sample. In essence, IMLE minimises the distance between each data point and its nearest generated sample, guaranteeing comprehensive mode coverage and maintaining stable training, even with smaller datasets. This makes IMLE a compelling alternative to diffusion models for developing efficient and faster behaviour cloning methods suitable for real-world robotics. 

In this work, we propose IMLE Policy, a novel extension of Implicit Maximum Likelihood Estimation tailored for conditional behaviour cloning.
We leverage a particular instantiation of IMLE that utilises rejection sampling in the training objective \cite{rsimle}, which further improves sample efficiency while effectively enabling the model to sample all possible modes during inference. IMLE Policy offers several key advantages that make it particularly promising for real-world robotic systems:

\begin{itemize}
    \item \textbf{Expressive multi-modal action distributions:} 
    By optimising the policy to generate samples close to every possible data point, IMLE Policy by nature does not drop modes, enabling highly expressive multi-modal action generations for a given state (Figure \ref{fig:main} (Bottom)).

    \item \textbf{Sample efficiency:} IMLE Policy can accurately model complex multi-modal distributions using 38\% less data on average than Diffusion Policy and single-step Flow-Matching to achieve similar performance (Figure \ref{fig:main} (Top)).

    \item \textbf{Fast inference:} IMLE Policy can generate multi-modal actions using only a single forward pass -- a 97.3\% increase in inference speed when compared to vanilla Diffusion Policy without collapsing modes when compared to single step Flow Matching.
    
\end{itemize}

We evaluate IMLE Policy across diverse simulation benchmarks and multi-modal real-world tasks, demonstrating its effectiveness as an alternative to existing generative model-based behaviour cloning approaches. IMLE Policy meets the critical \textit{desiderata} for real-world robotics: \textbf{sample efficiency, computational speed, and expressiveness}. \textit{(1) IMLE Policy outperforms state-of-the-art baselines in low-data regimes, while matching performance when more data is available}, learning performant real-world policies from as few as \textbf{17 demonstrations}. \textit{(2) IMLE Policy captures multi-modal action distributions in a single step}, generating \textbf{diverse} trajectories across varying conditions, avoiding mode collapse seen in other single-step methods. \textit{(3) IMLE Policy enables fast real-time inference}, reducing latency by \textbf{97.3\%} compared to iterative diffusion models while maintaining competitive task success rates. We further analyse sample efficiency, multi-modal expressivity, and key design choices to understand IMLE Policy’s strengths. All code, datasets, and training details will be released to ensure reproducibility and support further research. We provide supplementary videos to demonstrate the real world performance of our system when compared to the baselines.

\begin{figure*}[t]
    \centering
    \includegraphics[width=1.0\linewidth]{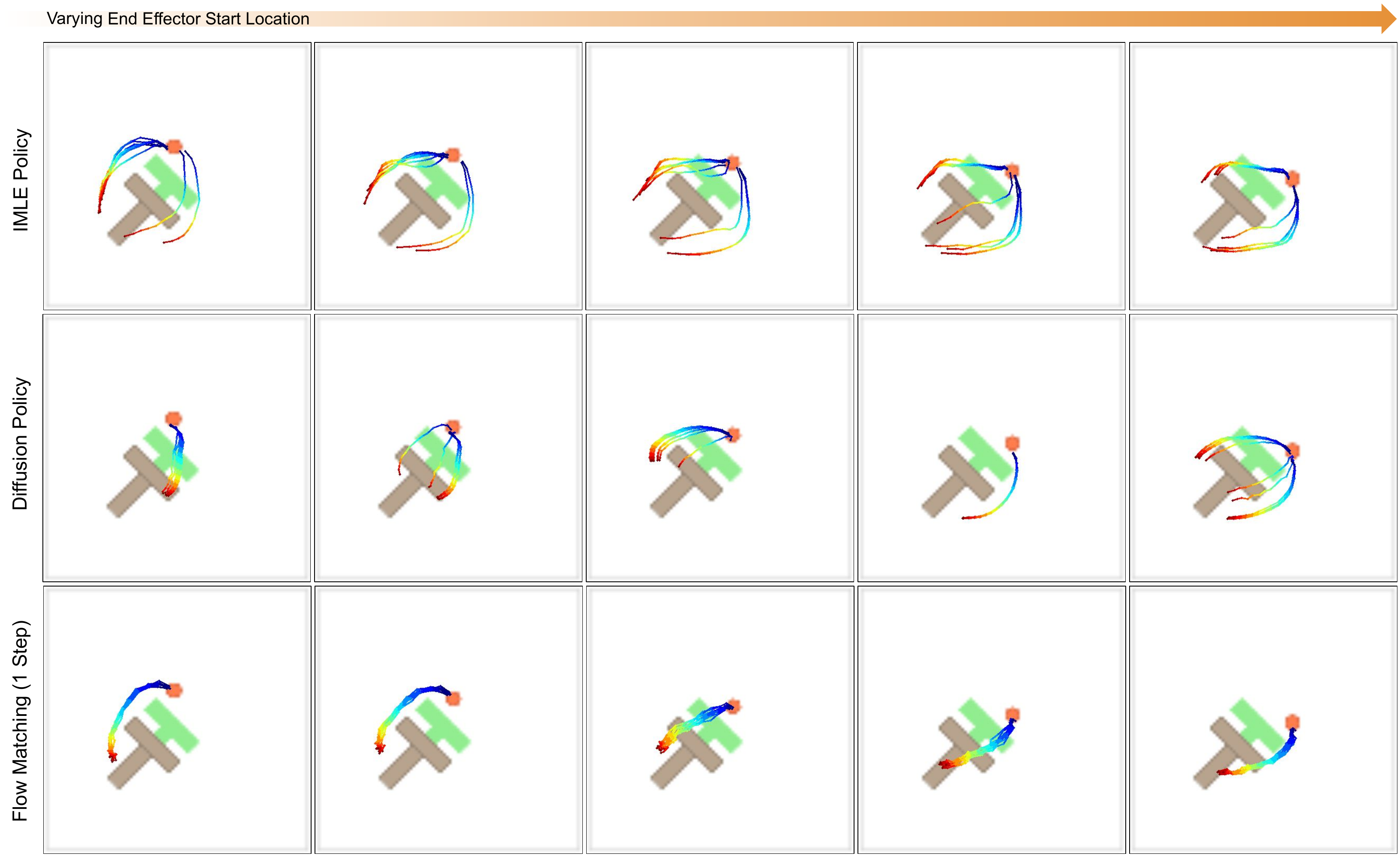}
    \caption{\textbf{Qualitative Analysis of Mode Capturing 
    Performance.} Illustrating the trajectories generated by different methods (rows) for varying initial end-effector positions (columns) in the Push-T task. The goal is to push the T-shaped block from its initial position to align with the target (green). At the top of the T, the dataset exhibits multi-modal behaviour: in the centre, demonstrations are evenly split between pushing left and right, while towards the edges, the majority of demonstrations push towards the closer side, with only a few moving in the opposite direction. IMLE Policy successfully captures all modes without collapsing or biasing, maintaining trajectory diversity even in underrepresented modes. In contrast, Diffusion Policy tends to bias towards one mode when close to an edge, while Flow Matching collapses modes, producing averaged and unimodal trajectories.}
    \label{fig:pusht_multimodality}
\end{figure*}

\section{Related Work}

We organise our discussion of prior work based on three key behaviour cloning desiderata for real-world robotics: capturing multi-modal behaviour, achieving sample efficiency, and enabling fast inference. While existing works address each of these aspects individually, relatively few approaches attempt to satisfy all three simultaneously.

\subsection{Multi-Modality in Behaviour Cloning}

Behaviour cloning has been widely explored as a means of enabling robots to learn from human demonstrations, achieving success across various manipulation tasks \cite{zhang2018deep, florence2019self, mandlekar2020learning, mandlekar2020iris, zeng2021transporter, rahmatizadeh2018vision, avigal2022speedfolding, atkeson1997robot, argall2009survey, ravichandar2020recent, chen2024diffusion, chiDiffusionPolicyVisuomotor2023b, Chen-RSS-24, Chi-RSS-24}. Capturing the multi-modal nature of human demonstrations is a fundamental challenge in behaviour cloning, and generative models have been extensively applied to address this issue. Methods such as Conditional Variational Autoencoders (CVAEs) \cite{lynch2020learning, zhaoLearningFineGrainedBimanual2023}, Energy-Based Models (EBMs) \cite{ibc}, and Vector Quantization \cite{leebehavior} have been used to model multi-modal action distributions, though each comes with trade-offs such as mode collapse or complex multi-stage training.

Recently, de-noising diffusion models have emerged as a dominant approach due to their stable training dynamics and strong performance in multi-modal imitation learning \cite{chen2024diffusion, chiDiffusionPolicyVisuomotor2023b, Chen-RSS-24, Chi-RSS-24, liu2024rdt}. However, these models require an iterative sampling process during inference, making them computationally expensive. 

An alternative class of generative models, IMLE (Implicit Maximum Likelihood Estimation), has remained largely unexplored in robotics despite its effectiveness in learning expressive distributions. In this paper, we introduce an IMLE-based approach for behaviour cloning, demonstrating its ability to efficiently model complex multi-modal action distributions in a single inference step, offering a simpler and more efficient alternative to existing diffusion-based methods.

\subsection{Sample Efficiency of Behaviour Cloning}
Despite the widespread success of diffusion models in robotics, these methods typically require large datasets for effective training \cite{zhao2024aloha}, which is a significant limitation in imitation learning, where expert demonstrations are expensive to collect. Improving the sample efficiency of behaviour cloning remains an active research area. 

Current research directions for improving sample efficiency involve optimising input representations. For example, 3D Diffusion Policy utilises point clouds to improve generalisation in 3D space \cite{ze20243d}, while approaches leveraging SO(2) and SIM(3) equivariances improve learning efficiency in tasks with structured transformations \cite{wangequivariant, yang2024equibot}. Affordance-centric representations have also been explored to improve generalisation from limited demonstrations \cite{rana2024affordance}. 

Our approach is orthogonal to these representation-based methods, as it focuses on improving the core learning algorithm itself to enhance sample efficiency. This allows IMLE Policy to be complementary to existing strategies, meaning it can be combined with improved input representations for even greater efficiency gains. Additionally, we conduct a comprehensive study on sample efficiency, explicitly examining the relationship between dataset size and performance. While dataset size is often an overlooked factor in behaviour cloning research, we highlight its direct impact on policy performance, providing insights into how different behaviour cloning methods scale with available data.

\subsection{Inference Speed and Behaviour Cloning}
A key limitation of most state-of-the-art behaviour cloning approaches is their multi-step inference process, which requires iterative de-noising or auto-regressive steps to generate actions. This significantly increases computational cost during inference, limiting real-time applicability in robotics. Several approaches aim to improve upon this. Consistency Models distill multi-step diffusion policies into single-step policies while maintaining performance \cite{prasad2024consistency, song2023consistency}. Streaming Diffusion modifies the denoising process to allow earlier actions to require fewer de-noising steps, speeding up inference \cite{høeg2024streamingdiffusionpolicyfast}. Flow Matching provides an alternative continuous-time generative modelling framework with more efficient probability paths, reducing inference steps \cite{lipmanflow, liuflow}, though in practice, it still requires multiple steps to prevent mode collapse when applied to behaviour cloning \cite{hu2024adaflow, black2024pi_0, zhang2024affordance}.

While these methods aim to speed up inference, they either require additional distillation steps or do not fully eliminate iterative sampling. In contrast, our approach natively enables single-step inference while still capturing complex, multi-modal action distributions. This allows IMLE Policy to perform fast, real-time inference without requiring multi-stage training or distillation, making it a promising alternative for computationally constrained settings.

\section{Method}

\begin{figure*}[t]
    \centering
    \includegraphics[width=1.0\linewidth]{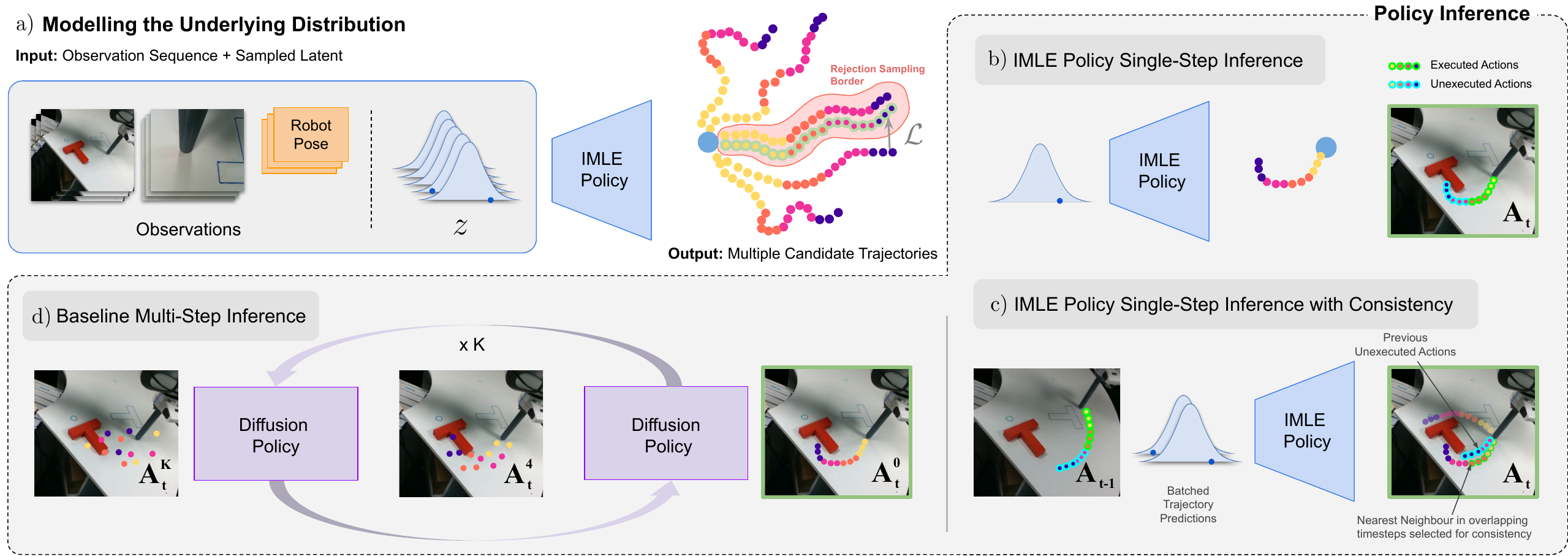}
    \caption{\textbf{IMLE Policy Overview}  a) Training: The policy takes in a sequence of past observations $\mathcal{O}$ and $m$ sampled latents $\textbf{z}$ for which the policy generates $m$ sequences of predicted actions $\mathcal{A}$. Generated trajectories that lie within the rejection sampling threshold $\epsilon$ are rejected. From the remaining trajectories, the nearest-neighbour to the ground truth trajectory is selected for training. We minimise the the distance between this trajectory and the ground truth trajectory to optimise the policy. As the loss focuses on each data sample, it ensures that all modes are captured even from small datasets. b) When compared to baselines with similar multi-modal capturing capabilities, IMLE can generate actions with a single inference step as opposed to multi-step de-noising processes. c) For highly multi-modal tasks, we enhance the performance of IMLE Policy by introducing a simple inference procedure to induce consistency upon mode choice based on a nearest-neighbour search over batch-generated action proposals with the previously executed action sequence.}
    \label{fig:enter-label}
\end{figure*}

\subsection{Background}
In the context of generative modelling, the aim is to learn the probability distribution of samples $p(x)$, which would allow us to then synthesise new samples via the trained model. We can represent the generator as a function $T_\theta : Z \rightarrow X$ that transforms samples from the latent space $Z$ to the space of target samples $X$, implemented as a neural network with parameters $\theta$. Such models have been historically trained via a generative adversarial objective (as in GANs), but this approach is prone to mode collapse, where only some modes of the target distribution are modelled.

A more recent approach, Implicit Maximum Likelihood Estimation (IMLE) has been introduced, providing an alternative training objective for the generator that avoids mode collapse \cite{imle}. The IMLE objective ensures that each training sample is well represented by the generator with samples generated nearby, alleviating the mode collapse issue. 

The IMLE training objective \cite{imle} is written as:
\begin{align}
    \theta_{\text{IMLE}}
    &= \argmin_{\theta} \mathbb{E}_{z_1,...,z_m \sim \mathcal{N}(0, I)}  \left [\sum_{i=1}^n \operatorname{min}\limits_{j\in [m]} d\left( {x}_i, T_\theta({z}_j)\right) \right ],
\label{eqn:imle}
\end{align}

where $d(\cdot,\cdot)$ is a distance metric, $n$ is the number of data samples, and $m$ is the number of generated samples. 

During training, $m$ number of samples $z_j$ are drawn from the latent prior distribution, a standard Gaussian distribution. These are transformed by the generator  $T_\theta$ into synthesised samples. For each training sample, the nearest synthetic sample is selected according to the distance metric $d(\cdot,\cdot)$. While this objective is effective for avoiding mode collapse, the selection procedure results in certain latent samples being rarely selected, even if they have a high likelihood under the latent prior distribution.

Rejection Sampling IMLE (RS-IMLE) \cite{rsimle} alleviates this issue by rejecting samples from the selection process if $d(x_i,T_\theta(z)) < \epsilon$, meaning they are too close to the training data sample. The remaining samples are then used in the IMLE training objective as before. Intuitively, this prevents the selection process from repeatedly selecting similar samples after they have already converged to fitting the data sample with some accuracy, defined by the parameter $\epsilon$.

\begin{algorithm}[t]
\SetAlgoLined
\SetKwInOut{Input}{Input}
\SetKwInOut{Output}{Output}
\caption{\textbf{IMLE Policy Training}}
\label{alg:training}

\Input{Training dataset $\mathcal{D} = \{(\mathcal{O}_i, \mathcal{A}_i)\}_{i=1}^n$, number of latents $m$, rejection threshold $\epsilon$, distance metric $d(\cdot, \cdot)$, generator $\pi_\theta(z, y)$.}
\Output{Trained policy $\pi_\theta$.}

\ForEach{$(\mathcal{O}_i, \mathcal{A}_i) \in \mathcal{D}$}{
    Sample $m$ latent vectors $\{z_j\}_{j=1}^m \sim \mathcal{N}(0, I)$\;
    Generate $m$ trajectories $\{\mathcal{A}_j\}_{j=1}^m$, where $\mathcal{A}_j = \pi_\theta(z_j, \mathcal{O}_i)$\;
    Compute distances $d_j = d(\mathcal{A}_i, \mathcal{A}_j)$ for $j \in [m]$\;
    Filter valid trajectories $\mathcal{V} = \{j \in [m] : d_j \geq \epsilon\}$\;
    Select nearest trajectory $j^* = \argmin_{j \in \mathcal{V}} d_j$\;
    Update $\theta$ to minimise $d(\mathcal{A}_i, \mathcal{A}_{j^*})$\;
}
\Return{$\pi_\theta$}
\end{algorithm}

\begin{algorithm}[t]
\SetAlgoLined
\SetKwInOut{Input}{Input}
\SetKwInOut{Output}{Output}
\caption{\textbf{Inference with Temporal Consistency in IMLE-Policy}}
\label{alg:inference_temporal_consistency}

\Input{Policy $\pi_\theta$, observation $\mathcal{O}$, number of latents $m$, distance metric $d(\cdot, \cdot)$, previously executed trajectory $\mathcal{A}_{\text{prev}[\mathcal{T}_{a}:\mathcal{T}_{p}]}$.}
\Output{Action sequence $\mathcal{A}_{[0:\mathcal{T}_{p}]}$.}

Sample $m$ latent vectors $\{z^j\}_{j=1}^m \sim \mathcal{N}(0, I)$\;
Batch generate $m$ trajectories $\{\mathcal{A}^j\}_{j=1}^m$, where $\mathcal{A}^j = \pi_\theta(z^j, \mathcal{O})$\;

Compute overlaps $o^j = d(\mathcal{A}_{\text{prev}[\mathcal{T}_{a}:\mathcal{T}_{p}]}, \mathcal{A}^j_{[0:\mathcal{T}_{a}]})$ for $j \in [m]$\;
Select trajectory $j^* = \argmin_{j \in [m]} o^j$\;

\Return{$\mathcal{A}_{j^*}$}
\end{algorithm}

\subsection{Conditional RS-IMLE}
We extend the RS-IMLE framework \cite{rsimle}, initially developed for unconditional image generation, to the conditional setting, enabling its application to behavior cloning. In this case, our generator is augmented to include an additional conditioning variable $y$, and instead of the $m$ samples shared between all data points, we now select $m$ samples to synthesise \textit{for each} data point, each sharing the conditioning value of that point.


Formally, for each training sample ($x_i, y_i$), we draw $m$ latent vectors $\{z_j\}_{j=1}^m$ from the standard Gaussian prior. These latent vectors are then transformed and filtered based on the rejection sampling criterion, resulting in a set of valid latent vectors $\mathcal{V}_i$:

\begin{align}
\mathcal{V}_i &= \{{z}_j \sim \mathcal{N}(0, I) : d({x}_i, T_\theta({z}_j, y_i)) \geq \epsilon, j=1,...,m\}
\end{align}

The valid latent vectors in $\mathcal{V}_i$ are transformed by the conditional generator $T_\theta({z}, {y}_i)$ to produce candidates, all conditioned on $y_i$. The nearest neighbour selection is then performed within this set of candidates specific to ${x}_i$, according to:  $\operatorname{min}\limits_{{z} \in \mathcal{V}_i} d\left( {x}_i, T_\theta({z}, y_i\right)) $. This per-example conditioning ensures that all candidates generated for a given training example share the same conditional context $y_i$, allowing the model to learn the conditional relationships present in the training data.

The overall objective for the Conditional RS-IMLE is:
\begin{align}
\label{eqn:crsimle}
\theta_{\text{C-RS-IMLE}} = \argmin_{\theta} \sum_{i=1}^n \mathbb{E}_{\mathcal{V}_i} \left[ \min_{{z} \in \mathcal{V}_i} d\left( {x}_i, T_\theta({z}, y_i)\right) \right],
\end{align}

\noindent where $\theta_{\text{C-RS-IMLE}}$ represents the optimal parameters of the model $T_\theta$ under the Conditional RS-IMLE objective. These parameters minimize the sum of expected minimum distances between each real data point ${x}_i$ and the generated outputs $T_\theta({z}, {y}_i)$ over the valid latent vectors ${z}$ in the corresponding set $\mathcal{V}_i$.

\subsection{IMLE Policy}
With our conditional variant of RS-IMLE, we can now apply this training objective (Eq. \ref{eqn:crsimle}) to behaviour cloning, where we formulate our policy as a generator $\pi(z,\mathbf{o}) \mapsto \mathbf{a}$, where our conditioning variable $\mathbf{o}$ takes the form of image and robot state information and $\mathbf{a}$ represents the generated actions. For the distance function $d(\cdot,\cdot)$ we use Euclidean distance. We utilise the same action generation procedure as \cite{chi2023diffusion}, generating a sequence of actions which we can utilise for closed-loop receding-horizon control. At each time step $t$, the policy processes the latest $T_o$ steps of observation data, $\mathcal{O}_t$, as input and predicts $T_p$ steps of future actions. Out of these, $T_a$ steps are executed on the robot before re-planning occurs. Here, $T_o$ is referred to as the observation horizon, $T_p$ as the action prediction horizon, and $T_a$ as the action execution horizon.
Although we use the same UNet architecture as Diffusion in this work, the approach is generally applicable to other architectures such as transformers. 
We summarise the IMLE Policy training algorithm in Algorithm \ref{alg:training}


\subsection{Temporal Consistency}

IMLE Policy effectively models multi-modal action distributions; however, since each inference step is independent, the policy may exhibit \textit{mode-switching} at decision points where multiple valid behaviour exist. This can lead to oscillatory behaviour, particularly in long-horizon tasks where smooth and consistent execution is necessary. To mitigate this, we introduce a \textit{batched trajectory selection mechanism} that enforces temporal consistency while maintaining the policy’s multi-modal expressivity during inference.  

At each time step \( t \), we generate a batch of \( m \) latent-conditioned trajectories,  
\begin{align}
\mathcal{A}^j = \pi_\theta(z^j, \mathcal{O}_t), \quad \text{for } j \in [m],
\end{align}
\noindent where \( z^j \sim \mathcal{N}(0, I) \) are sampled latent vectors, and \( \mathcal{A}^j \) represents a candidate trajectory. To maintain temporal consistency, we select the trajectory that minimises deviation from the previously generated trajectory over the overlapping horizon:

\begin{align}
j^* = \argmin_{j \in [m]} d(\mathcal{A}_{\text{prev}[\mathcal{T}_{a}:\mathcal{T}_{p}]}, \mathcal{A}^j_{[0:\mathcal{T}_{a}]}),
\end{align}

\noindent where \( d(\cdot, \cdot) \) is a distance metric, \( \mathcal{A}_{\text{prev}[\mathcal{T}_{a}:\mathcal{T}_{p}]} \) represents the unexecuted segment of the previously generated trajectory and  \( \mathcal{A}^j_{[0:\mathcal{T}_{a}]} \) represents the executable $T_a$ actions from the generated action sequence. As in training, we use Euclidean distance here for $d(\cdot,\cdot)$. The first $T_a$ steps of the selected trajectory \( \mathcal{A}^{j^*} \) are then executed at timestep \( t \).  

To prevent the policy from overcommitting to a suboptimal trajectory, we introduce a periodic trajectory reset. Every \( C \) steps, a new trajectory is randomly selected from the candidate batch:
\begin{align}
j^* = \text{Uniform}([m]).
\end{align}

\noindent This ensures adaptability while preventing long-term commitment to potentially poor behaviours.  

Notably, mode-switching is a challenge inherent to any model that accurately captures multi-modal behaviour, as the policy must balance between expressivity and trajectory consistency. Prior works have explored alternative strategies such as bidirectional decoding and streaming diffusion   \cite{liu2024bidirectional,høeg2024streamingdiffusionpolicyfast}. However, since IMLE Policy is a single-step generative model, our batch-based selection strategy is straightforward and computationally efficient and can run at $>100$ Hz.
We summarise inference using IMLE policy with consistency in Algorithm \ref{alg:inference_temporal_consistency}.

\section{Evaluation}

\begin{table}[!b]
\centering
\setlength{\tabcolsep}{4pt} 
\renewcommand{\arraystretch}{1.2} 
\begin{tabular}{@{}lccccccc@{}}
\toprule
Task & \#Rob & \#Obj & ActD & \#Demo  & Steps & Img? & HiPrec \\ 
\midrule
\multicolumn{8}{l}{\cellcolor[HTML]{C0C0C0}Simulation Benchmark} \\
Lift       & 1 & 1 & 7  & 200  & 400 & Yes & No  \\
Can        & 1 & 1 & 7  & 200 & 400 & Yes & No  \\
Square     & 1 & 1 & 7  & 200  & 400 & Yes & Yes \\
Transport  & 2 & 3 & 14 & 200  & 700 & Yes & No  \\
ToolHang   & 1 & 2 & 7  & 200   & 700 & Yes & Yes \\
Push-T     & 1 & 1 & 2  & 200    & 300 & Yes & Yes \\
UR3 Block Push  & 1 & 2 & 2  & 1000    & 350 & No  & No  \\
Kitchen    & 1 & 7 & 9  & 18   & 280 & No  & No  \\ 
\midrule
\multicolumn{8}{l}{\cellcolor[HTML]{C0C0C0}Realworld Benchmark} \\
Push-T             & 1 & 1 & 2 & 35    & 600 & Yes & Yes \\
Shoe Rack          & 1 & 3 & 7 & 35   & 800 & Yes & Yes  \\ 
\bottomrule
\end{tabular}
\caption{\textbf{Tasks Summary.} \#Rob: number of robots, \#Obj: number of objects, ActD: action dimension, \#Demo: number of demonstrations, Steps: max number of rollout steps, HiPrec: whether the task has a high precision requirement.}
\label{tab:sim_env_info}
\end{table}

\begin{figure*}
    \centering
    \includegraphics[width=1.0\linewidth]{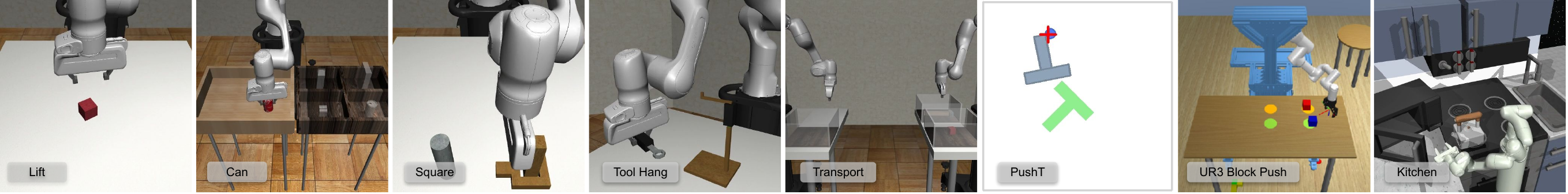}
    \caption{\textbf{Simulation Environments.} Visualisation of the environments used to evaluate IMLE Policy. The first 5 tasks are part of the Robomimic benchmark \cite{robomimic} and the other environments include Push-T \cite{chi2023diffusion}, UR3 BlockPush \cite{kim2022automating} and Franka Kitchen \cite{gupta2019relay}.}
    \label{fig:tasks}
\end{figure*}

We evaluate IMLE Policy to understand how effective it is in modelling behaviours across different datasets and environments when compared to prior state-of-the-art methods. We focus our evaluation on the core requirements for real-world robotics applications particularly: sample efficiency, multi-modal expressivity and inference speed. Concretely, we seek to answer the following questions through our experiments:

\begin{enumerate}
    \item How well does IMLE Policy perform on the respective benchmarks for behaviour cloning as a single-step generative model?
    \item How well does IMLE Policy perform when under data constraints?
    \item How well does IMLE capture multiple modes seen in demonstrations?
    \item Does IMLE Policy scale beyond simulation environments?
    \item What hyper-parameters of IMLE Policy make the most impact on its performance?
\end{enumerate}

We systematically evaluate IMLE Policy to answer the above questions through a series of experiments across both simulated and real-world environments. For all experiments, we evaluate the following policies: 

\begin{itemize}
\item \noindent\textit{IMLE Policy: } Our proposed algorithm given in Algorithm \ref{alg:inference_temporal_consistency} which leverages temporal consistency during inference.
\item\textit{IMLE Policy (w/out consistency): } Our proposed algorithm without temporal consistency during inference.
\item\textit{Diffusion Policy: } Vanilla Diffusion Policy proposed by \citet{chi2023diffusion} trained using DDPM and 100 denoising steps
\item\textit{Flow Matching (1-step): } We modify the Diffusion Policy implementation from \citet{chi2023diffusion} to utilise the Flow Matching objective and evaluate against the 1-step setting as a comparative baseline to the 1-step performance of our method.
\end{itemize}

\begin{table*}[]
\centering
\resizebox{\textwidth}{!}{%
\begin{tabular}{@{}l
>{\columncolor[HTML]{C0C0C0}}c 
>{\columncolor[HTML]{C0C0C0}}c 
>{\columncolor[HTML]{C0C0C0}}c 
>{\columncolor[HTML]{FFFFFF}}c 
>{\columncolor[HTML]{FFFFFF}}c 
>{\columncolor[HTML]{FFFFFF}}c 
>{\columncolor[HTML]{FFFFFF}}c 
>{\columncolor[HTML]{FFFFFF}}c @{}}
\toprule
 &
  \multicolumn{3}{c}{\cellcolor[HTML]{C0C0C0}High Multimodality} &
  \multicolumn{5}{c}{\cellcolor[HTML]{FFFFFF}Low Multimodality} \\ \midrule
\multicolumn{1}{c}{} &
  Push-T &
  UR3 Block Push&
  Kitchen &
  Lift &
  Can &
  Square &
  Tool Hang &
  Transport \\ \midrule
\multicolumn{9}{c}{\cellcolor[HTML]{EFEFEF}\textbf{100\% Dataset}} \\
\cellcolor[HTML]{FFFFFF}\textbf{IMLE Policy} &
  \textbf{0.59/0.54} &
  0.82/0.73 &
  \textbf{0.11/0.09} &
  \textbf{1.00/1.00} &
  0.98/0.97 &
  0.82/0.80 &
  0.81/0.69 &
  0.90/0.80 \\
\cellcolor[HTML]{FFFFFF}\textbf{IMLE Policy (w/out consistency)} &
  0.56/0.53 &
  \textbf{0.90/0.77} &
  0.07/0.04 &
  \textbf{1.00/1.00} &
  0.96/0.95 &
  \textbf{0.86/0.82} &
  0.74/0.52 &
  0.92/0.80 \\
\cellcolor[HTML]{FFFFFF}Diffusion Policy &
  0.57/0.52 &
  0.74/0.73 &
  0.10/0.06 &
  \textbf{1.00/1.00} &
  \textbf{0.99/0.98} &
  0.84/0.84 &
  \textbf{0.84/0.81} &
  0.92/0.90 \\
\cellcolor[HTML]{FFFFFF}Flow Matching Policy &
  0.36/0.34 &
  0.87/0.81 &
  0.09/0.05 &
  \textbf{1.00/1.00} &
  0.96/0.96 &
  0.82/0.79 &
  0.78/0.68 &
  \textbf{0.93/0.87} \\ 
\multicolumn{9}{c}{\cellcolor[HTML]{EFEFEF}\textbf{20 Demos}} \\
\cellcolor[HTML]{FFFFFF}\textbf{IMLE Policy} &
  \textbf{0.10/0.07} &
  \textbf{0.34/0.32} &
  - &
  \textbf{1.00/0.99} &
  \textbf{0.50/0.41} &
  \textbf{0.18/0.14} &
  \textbf{0.03/0.00} &
  \textbf{0.28/0.20} \\
\cellcolor[HTML]{FFFFFF}\textbf{IMLE Policy (w/out consistency)} &
  0.05/0.03 &
  0.34/0.31 &
  - &
  1.00/0.98 &
  0.44/0.40 &
  0.11/0.09 &
  0.02/0.00 &
  0.27/0.22 \\
\cellcolor[HTML]{FFFFFF}Diffusion Policy &
  0.03/0.03 &
  0.26/0.21 &
  - &
  \textbf{1.00/0.99} &
  0.47/0.47 &
  0.16/0.16 &
  0.00/0.00 &
  0.27/0.24 \\
\cellcolor[HTML]{FFFFFF}Flow Matching Policy &
  0.012/0.00 &
  0.21/0.14 &
  - &
  \textbf{1.00/0.99} &
  0.48/0.39 &
  0.16/0.08 &
  0.02/0.00 &
  0.26/0.15 \\ \bottomrule
\end{tabular}%
}
\caption{\textbf{Behaviour Cloning Benchmark} We present success rates in the format of (max success rate) / (average of last 3 checkpoints), with each averaged across 3 training seeds and 50 different environment initial conditions.}
\label{tab:sim_results}
\end{table*}

\subsection{Simulation Environments}

IMLE Policy is evaluated across 8 different simulation tasks (Figure \ref{fig:tasks}) from 4 benchmark environments \cite{chi2023diffusion,robomimic,kim2022automating,gupta2019relay}. We categorise our sets of tasks based on the level of multi-modality exhibited by the dataset with the \textit{high multi-modality} tasks exhibiting the highest variance across the demonstrations. A high-level description of each task is provided below and specific details are summarised in Table \ref{tab:sim_env_info}.

\subsubsection{Push-T } We utilise the variant of this task provided by \citet{chi2023diffusion} which involves pushing a T-shaped block (gray) to a designated target position (red) using a circular end-effector (blue). Variation is introduced through randomized initial positions of the T block and the end-effector. Successfully completing the task requires leveraging complex, contact-rich object dynamics to precisely push the T block using point contacts to sufficiently overlap with the target.

\subsubsection{Robomimic }A large-scale robotic manipulation benchmark \cite{robomimic}  designed for studying imitation learning and offline reinforcement learning (RL). It includes five tasks and we utilise the Proficient Human (PH) teleoperated demonstration dataset for all evaluations. Successful completion of these tasks occurs when each of the objects is placed in the target locations.

\subsubsection{UR3 Block Push } In this task, a UR3 robot pushes two blocks to goal circles on the opposite side of the table \cite{kim2022automating}. The dataset exhibits multimodal behaviour, as either block can be moved first. Success is evaluated based on whether each block reaches the goal. 

\subsubsection{Franka Kitchen } This environment, based on the Franka Kitchen manipulation task \cite{gupta2019relay}, uses a Franka Panda robotic arm with a 7-dimensional action space and 18 successful human-collected demonstrations. It features seven possible tasks, with each trajectory completing four tasks in varying sequences making it highly multimodal. Successful completion in this environment is considered when a series of 4 unique tasks are completed sequentially.

\subsection{Real World Tasks} \label{sec:realworldtasks}

We additionally evaluate IMLE Policy on two real robot manipulation tasks to demonstrate its capability of learning manipulation tasks in the real world. To investigate real-world sample-efficiency, we only collect 35 demonstrations and train on 35 and 17  demonstration subsets.

\subsubsection{Real World Push-T}
\begin{figure}
    \centering
    \includegraphics[width=1.0\linewidth]{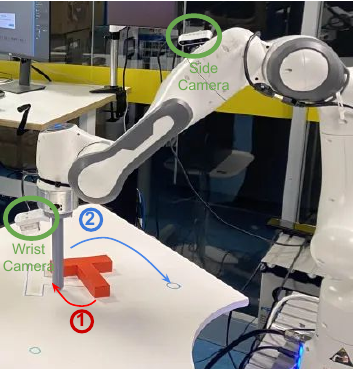}
    \caption{\textbf{Real World Push-T Task Setup.} Hardware setup and illustration of the task. The robot needs to \ding{172} precisely push the T-Shaped block into the target region, and \ding{173} move the end-effector to the end-zone.}
    \label{fig:real_pusht}
\end{figure}

This task is a real world variant of the simulated Push-T task, with a 3D printed T block and a cylindrical robot end-effector being controlled on a 2D plane. We use 2 camera views, one on the wrist of the robot, and the other a side view scene camera. We record a successful trial when the robot has successfully placed the T-Block to sufficiently overlap with the target and then return to an end state within a 2 minute time frame, as shown in Figure \ref{fig:real_pusht}.

\subsubsection{Real World Shoe Racking Task}
\begin{figure}
    \centering
    \includegraphics[width=1.0\linewidth]{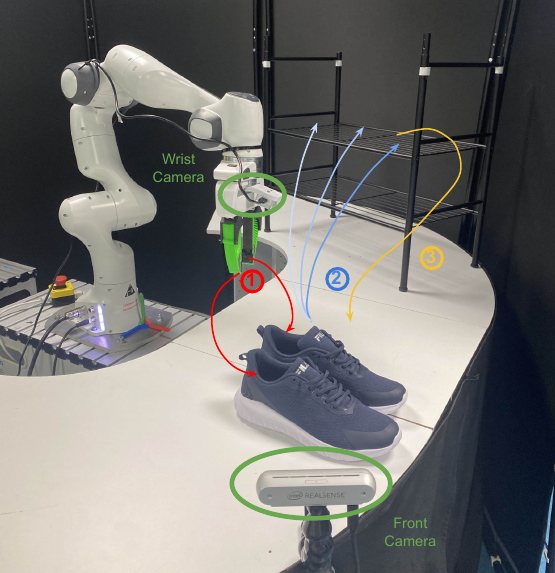}
    \caption{\textbf{Real World Shoe Racking Task Setup.} Hardware setup and illustration of the task. The robot needs to \ding{172} grasp either shoe and \ding{173} place it anywhere continuously along the top of the rack, this is repeated for the second shoe with the caveat that the shoe has to be placed either to the left or right with respect to the other shoe in order to be successful; finally \ding{174} the robot must return to its home position.}
    \label{fig:real_shoe}
\end{figure}

We evaluate our method on an additional multi-modal real-world shoe racking task, where the robot must pick and place two shoes and place them side-by-side on the shoe rack. Either shoe can be selected first, and the shoes can be placed anywhere on the rack, with demonstrations including a variety of these behaviours, making this task highly multi-modal. Wrist and scene cameras are used to provide full view of the task scene. A successful trial is recorded when both shoes are placed correctly side-by-side (i.e. left shoe on the left of the right shoe) on the rack and the robot has returned to its home position within a 1 minute time frame, as shown in Figure \ref{fig:real_shoe}

\begin{figure}[t]
    \centering
    \includegraphics[width=1.0\linewidth]{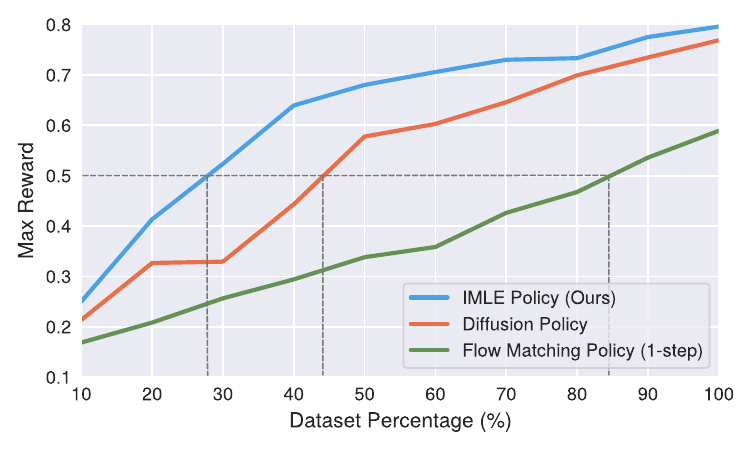}
    \caption{\textbf{Dataset Size Study.}  We evaluate how each of the methods performs when trained with different size subsets of the full dataset used in the Push-T benchmark task. The dashed line indicates that to achieve the same reward of 0.5 on the task, IMLE Policy requires less than 29\% of the data, while Diffusion Policy requires approximately 43\% while Flow Matching Policy requires over 80\% of the full dataset.}
    \label{fig:simpusht-sweep}
\end{figure}

\begin{figure}[t]
    \centering
    \includegraphics[width=1.0\linewidth]{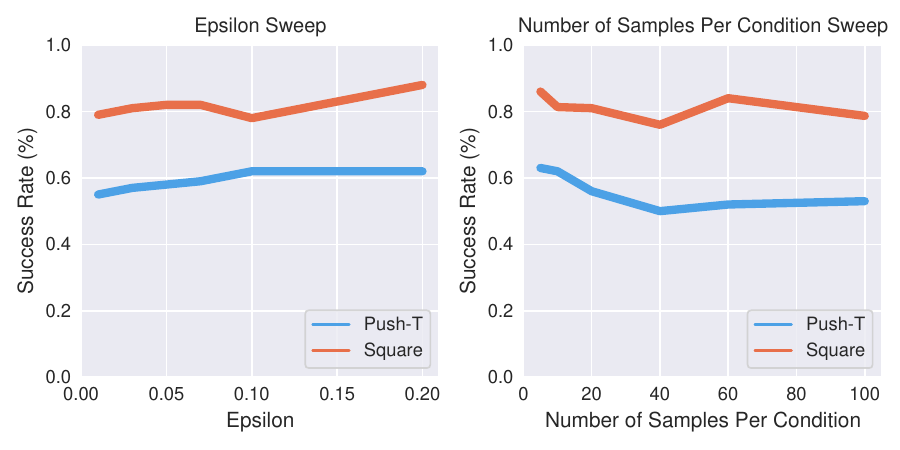}
    \caption{\textbf{Robustness to Hyper-parameters.} We evaluate the performance of IMLE Policy across varying different values for two key hyper-parameters used by the IMLE objective: rejection sampling radius $\epsilon$ and the number of samples used per condition. Note how the algorithm remain relatively stable across the entire range of values for both tasks.}
    \label{fig:hyper_sweep}
\end{figure}

\subsection{Experiments}
\subsubsection{Dataset Size Study}

We conduct a study to evaluate how well IMLE can learn under data constrains when compared to the baselines. We use the Push-T simulated task for this, dividing the full dataset provided by \citet{chi2023diffusion} in 10 subsets, starting with 10\% of the data up to 100\%. We train each method on these datasets for 1000 epochs across 3 seeds and report the average evaluation performance for each method in Figure \ref{fig:simpusht-sweep}. This experiment gives us a detailed view of how performance improves as data increases.

\subsubsection{Benchmark Evaluation}

We evaluate all methods across the full datasets provided by each respective benchmarks indicated as \textit{100\% Dataset} in Table \ref{tab:sim_results}. To understand how our policy operates in the low data regime, we additionally conduct an evaluation when all methods are trained on 20 randomly sampled demonstrations from the full dataset. These 20 demonstrations are kept consistent across all methods. While 20 demonstrations might not be enough data to learn to successfully complete each task, it allows us to gauge the relative sample efficiency of all methods. For each setting, we report the best success rate achieved by the method across 1000 training epochs. The success rates are reported every 50 epochs and are calculated as the average success across 50 different initialisations within that environment. We additionally report the average of the last 3 success rates from the run. All results are averaged across 3 seeds and are summarised in Table \ref{tab:sim_results}.

\subsubsection{Real World Validation} We evaluate all methods on the real-world Push-T and shoe racking tasks described in \ref{sec:realworldtasks}. Our evaluation protocol tests both sample efficiency and overall performance by training on 17 and 35 demonstrations for each task. For both tasks, we conduct 20 evaluation trials per method and report the success rates based on the completion criteria outlined in the environment descriptions. This evaluation framework allows us to assess how effectively each method can learn from limited real-world demonstrations while handling both precise manipulation requirements in Push-T and the multi-modal decision space in the shoe racking task. Results are summarised in Figure \ref{fig:realresults}

\subsubsection{Mode Capturing Ablation} In this qualitative ablation study we demonstrate how IMLE Policy can capture multiple modes even in states where certain modes appear less frequently in the dataset. We conduct this study in the Push-T simulation benchmark where we gradually sweep the location of the robot's end effector from one side of the T block's top edge to the other. The idea here is to capture points where the demonstration data exhibits a high level of multi-modality (at the centre) and less multi-modality (towards the corners) where majority of the demonstrations will bias towards one side of the T block when in one of these particular states. The key results are illustrated in Figure \ref{fig:pusht_multimodality}.

\subsubsection{Real World Inference Speed Evaluation} We evaluate the computational efficiency of our approach by measuring inference speed across different policy architectures for the shoe racking task. Using a standard Dell Precision 3680 i7 workstation equipped with an NVIDIA GeForce RTX 3090 GPU, we measure the average time required to generate a sequence of actions. For each method, we conduct 30 separate generations and report the average inference speed in Hertz. We summarise the results in Table \ref{tab:realworldspeed}. 


\subsubsection{Hyperparameter Ablation Study} Finally we evaluate how robust our algorithm is to the two key hyperparameters used by the IMLE Policy objective. The first parameter pertains to the rejection sampling radius, $\epsilon$ which determines how far candidate generations have to be away from a data point to be considered in the loss computation. The second parameter pertains to the number of samples we use per condition in order to conduct our nearest-neighbour search in the objective. For each hyper-parameter we sweep over a set of values for two different tasks from our simulation benchmark (Push-T and Square) and report the maximum success rate achieved by each when trained over 1000 epochs. The results are summarised in Figure \ref{fig:hyper_sweep}.

\begin{figure}[t]
    \centering
    \includegraphics[width=1.0\linewidth]{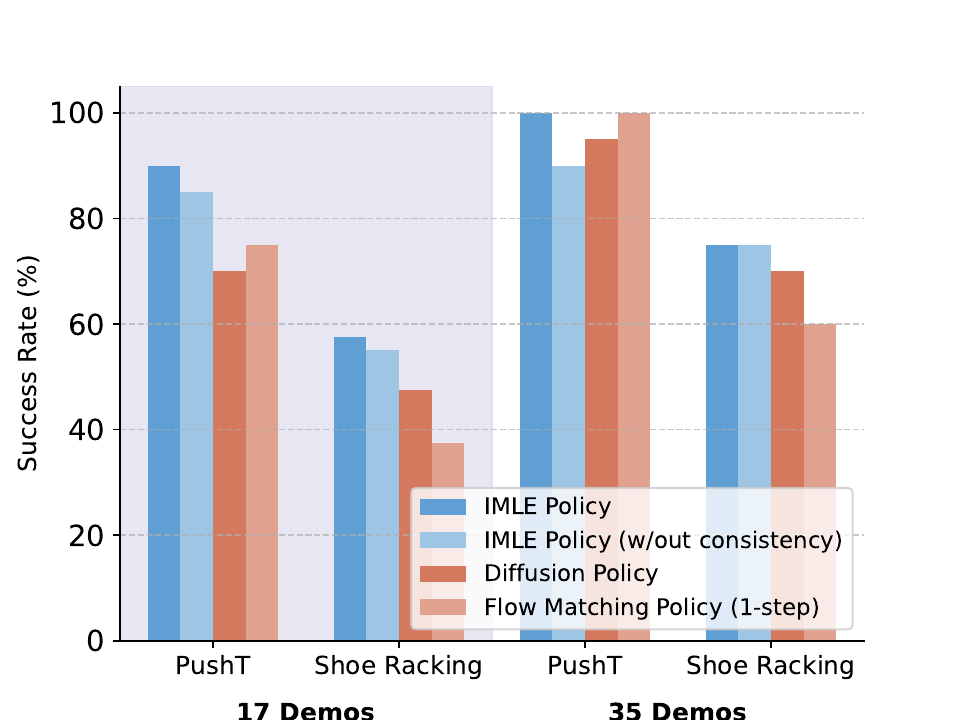}
    \caption{\textbf{Real World Task Results.} Across both tasks we train each method on both 17 (left) and 35 (right) demonstrations and report the average success rate achieved across 20 trials. We find that IMLE Policy consistently outperforms baselines in the low data regime, while maintaining competitive in the higher data regime.}
    \label{fig:realresults}
\end{figure}

\begin{table}[]
\centering
\resizebox{\columnwidth}{!}{%
\begin{tabular}{@{}lcccc@{}}
\toprule
 &
  \textbf{\begin{tabular}[c]{@{}c@{}}IMLE \\ Policy\end{tabular}} &
  \textbf{\begin{tabular}[c]{@{}c@{}}IMLE Policy \\ (w/out consistency)\end{tabular}} &
  \textbf{\begin{tabular}[c]{@{}c@{}}Diffusion \\ Policy\end{tabular}} &
  \textbf{\begin{tabular}[c]{@{}c@{}}Flow Matching \\ Policy (1-step)\end{tabular}} \\ \midrule
\textbf{Inference Speed (Hz)} &
  111 &
  123 &
  1.8 &
  110 \\ \bottomrule
\end{tabular}%
}
\caption{\textbf{Real World Inference Speed.} Using the two images from shoe racking task as input, we average the inference speed required to generate a sequence of actions across 30 generations.}
\label{tab:realworldspeed}  
\end{table}

\subsection{Key Findings}

\textbf{IMLE Policy can learn with significantly less data when compared to baselines.} The dataset size study shown in Figure \ref{fig:simpusht-sweep} demonstrates how IMLE Policy consistently achieves higher max rewards across all dataset percentages and a significant advantage in the lower data settings. Notably, IMLE Policy achieves a reward of 0.5 with less than 30\% of the data, while Diffusion Policy requires nearly twice as much data to reach the same performance. This underscores the ability of IMLE Policy to generalise effectively with minimal data. Furthermore, as the dataset size increases, IMLE Policy continues to maintain its advantage, achieving near-optimal performance faster, while converging to comparable performance to Diffusion Policy when the full dataset is available. 

\textbf{IMLE Policy can maximise the utility of only a few demonstrations, while scaling to larger datasets.} As shown in Table \ref{tab:sim_results}, across all benchmark environment tasks, IMLE Policy demonstrates a clear advantage, particularly in the challenging setting of learning from only 20 randomly selected demonstrations. Under this constrained scenario, IMLE Policy consistently outperforms all baseline methods,  achieving at least 8 more successful evaluation runs on average compared to the baselines. On the full dataset benchmark, IMLE Policy continues to demonstrate competitive or superior performance, outperforming the baselines in at least 5 of the 8 tasks. 

\textbf{IMLE Policy is highly multi-modal.} Figure \ref{fig:pusht_multimodality} shows that IMLE Policy maintains multi-modality across varying initial conditions of the end effector. Unlike Diffusion Policy, which biases towards majority modes, and Flow Matching Policy (one-step), which collapses modes and produces averaged trajectories, IMLE Policy captures diverse trajectories that reflect the full distribution of the demonstration data. This is particularly important in low-data regimes, where capturing all modes without overfitting to dominant behaviours enables efficient learning. Figure \ref{fig:main} further supports this, showing that IMLE Policy preserves multi-modality and accurately captures all modes, even with sparse or imbalanced data.

\textbf{IMLE Policy is robust to hyperparameter variations.} Figure \ref{fig:hyper_sweep} demonstrates that IMLE Policy consistently performs well across a wide range of hyperparameter settings, with stable success rates even as the rejection sampling threshold ($\epsilon$) and the number of samples per condition are varied. This robustness ensures that IMLE Policy adapts effectively without significant degradation in performance, making it practical for real-world applications where hyperparameter searches or fine-tuning may be infeasible. Notably, for the simulation benchmark evaluations, hyperparameters were not optimized per task but held constant throughout, yet IMLE Policy still outperformed baseline methods. 

\textbf{IMLE Policy is well suited for real world robotics.}
IMLE Policy demonstrates impressive traits that make it suitable for real-world robotics by achieving the best performance across both real-world visuomotor tasks as shown in Figure \ref{fig:realresults}. Even with as few as 17 demonstrations, IMLE Policy learns performant policies, outperforming both baseline methods by a significant margin. The robustness of IMLE Policy in low-data regimes makes it particularly valuable in real-world scenarios where collecting large datasets is often impractical. Additionally, as shown in Table \ref{tab:realworldspeed} IMLE Policy exhibits fast single-step inference, achieving up to 97.3\% faster inference speeds compared to vanilla Diffusion Policy, while either outperforming it or maintaining competitive performance. This efficiency enables real-time control and is well suited for robotics applications where computational resources can be limited. While its inference speed is similar to that of single-step Flow Matching we note that IMLE Policy outperforms Flow Matching, particularly in the low data regime and consistently across the simulation benchmarks. We provide videos demonstrating how our policy perform across both tasks when compared to the baselines and additional videos to demonstrate its robustness in the real world in the attached supplementary material.

\textbf{High multi-modality calls for temporal consistency.} While IMLE Policy demonstrates impressive multi-modal expressivity even from a limited number of demonstrations, a key empirical insight we identified was the tendency for the policy to occasionally switch between modes during execution in highly multi-modal tasks. This behaviour arises because the model is designed to represent all modes present in the data, and without explicit guidance, the policy may select a different mode at each decision point. This can result in inconsistent trajectories, particularly in tasks requiring sequential and coherent actions. Temporal consistency played an important role in ensuring that once a mode was selected, the policy adhered to it across subsequent timesteps. This not only stabilises and smooths executed trajectories but also prevents the policy from getting stuck between conflicting modes. As demonstrated in both the real-world and simulation experiments, temporal consistency significantly improves success rates, particularly in tasks where precise and sustained adherence to a trajectory is critical. The smooth and consistent nature of these trajectories can be viewed in the supplementary real world video attached to this submission.

\subsection{Implementation Details}
We base our implementation on the available Diffusion Policy training code \cite{chi2023diffusion}. As in Diffusion Policy, the model takes in an input noise vector, with the same dimensionality as the action sequence, however in our case we do not denoise this gradually into a clean trajectory, but rather use it as the latent sample space and directly output the trajectory. Therefore the model architecture is a 1D UNet, which is the same as Diffusion Policy, modified only to remove the timestep embedding. We use an action prediction horizon $T_p$ of $16$, and action execution horizon $T_a$ of $8$ and an observation horizon $T_o$ of 2. For all experiments we use fixed hyperparameters, with a value of $0.03$ for $\epsilon$, $20$ generated samples per datapoint. We train all simulation benchmark tasks for $1000$ epochs while all real world tasks were trained for a fixed time of 12 hours. For inference, when using consistency, we set the reset period $C$ to 10, which gives a good balance of consistency and responsiveness. Although we keep this value the same across tasks, note that it can be tuned for the time horizon and desired level of responsiveness of specific tasks. 


\section{Limitations}

While IMLE Policy demonstrates strong performance in multi-modal behaviour cloning, certain limitations remain. Its strong ability to capturing all modes makes it more sensitive to the quality of demonstrations, meaning inconsistencies or suboptimal behaviours in the dataset can negatively impact performance. Additionally, in highly multi-modal tasks, IMLE Policy may occasionally switch between modes during execution, necessitating consistency mechanisms such as temporal coherence; however, this can sometimes reinforce suboptimal behaviours if the selected future actions contain errors.
Recent concurrent work, such as the BID Diffusion Policy \cite{liu2024bidirectional}, introduces bidirectional decoding to bridge the tradeoff between long-term consistency and short-term reactivity by coupling past decisions with forward-looking strategies. Exploring similar decoding strategies could further enhance the robustness of IMLE Policy in multi-modal and stochastic environments. While IMLE Policy is computationally efficient at inference, training incurs some additional overhead due to nearest-neighbour (NN) searches. While we do not optimise this in our work, since we did not find the additional training cost to be too significant, the NN search could be improved using more efficient NN techniques as in prior IMLE research \cite{imle}. Furthermore, while our method outperforms baselines in most settings particularly in the low data regime, it does not consistently outperform Diffusion Policy in every benchmark task in the full dataset setting, though we note that Diffusion Policy still requires iterative inference, whereas IMLE Policy is significantly faster.

\section{Future Work}

IMLE Policy offers a promising foundation for advancing behaviour cloning in real-world robotics and presents several avenues for further research and applications. To facilitate exploration, we will release all code and datasets, encouraging the robotics and machine learning communities to investigate and build upon this work in a variety of domains.

\textbf{Exploring the use of IMLE Policy for low-cost robotics.} IMLE Policy’s efficient single-step inference and sample-efficiency presents an opportunity for adoption in low-cost, open-source robotics initiatives, such as LeRobot \cite{cadene2024lerobot}, where computational constraints limit the use of complex methods. Future research could investigate how IMLE Policy can be effectively leveraged in such settings to democratise access to high-quality robotic learning and enable broader adoption.

\textbf{Investigating reinforcement learning fine-tuning with IMLE Policy.} The simplicity of IMLE Policy's single-step inference makes it a promising candidate for reinforcement learning (RL) fine-tuning, potentially streamlining the adaptation of imitation-trained policies to new environments. Future work could explore how IMLE Policy simplifies RL finetuning pipelines compared to iterative methods like Diffusion Policy, which make RL fine-tuning difficult \cite{dppo}.

\textbf{Utilising IMLE Policy for downstream robotics applications.} IMLE Policy’s expressive multi-modal capabilities open the door to higher-level control processes, such as Model Predictive Control (MPC), where diverse trajectory proposals could enhance decision-making or potentially facilitate meaningful exploration in reinforcement learning.

\textbf{Scaling IMLE Policy to large and diverse datasets.} While IMLE Policy has shown strong performance in low-data regimes, an important area for exploration is its scalability to large-scale and diverse datasets, such as the Open-X Embodiment dataset \cite{o2024open}. Understanding its generalisation across multi-task scenarios would be an interesting area to explore.

\section{Conclusion} 
\label{sec:conclusion}

In this work, we introduced IMLE Policy, a novel imitation learning algorithm based on a conditional variant of RS-IMLE, designed to efficiently capture multi-modal action distributions while enabling fast, single-step inference. Through extensive experiments, we demonstrated state-of-the-art sample efficiency in both simulated and real-world robotic manipulation tasks, showing that IMLE Policy can learn effective policies from limited demonstrations. We conducted a thorough evaluation across varying dataset sizes, addressing an underexplored area in behaviour cloning research. Additionally, we proposed a variant that encourages temporal consistency without modifying the training procedure, enhancing execution stability in multi-modal settings. IMLE Policy exhibits promising characteristics for future research, including reinforcement learning fine-tuning, diverse behaviour generation for model predictive control (MPC) and RL exploration, and deployment in resource-constrained settings. Its efficiency and simplicity make it especially relevant for open-source robotics and real-world applications.

\section*{Acknowledgments}
The authors also acknowledge the ongoing support from the QUT Centre for Robotics. This work was partially supported by the Australian Government through the Australian Research Council's Discovery Projects funding scheme (Project DP220102398) and by an Amazon Research Award to Niko S\"underhauf. This work was also supported by the QUT Research Engineering Facility. 

\bibliographystyle{plainnat}
\bibliography{references}

\begin{thebibliography}{43}
\providecommand{\natexlab}[1]{#1}
\providecommand{\url}[1]{\texttt{#1}}
\expandafter\ifx\csname urlstyle\endcsname\relax
  \providecommand{\doi}[1]{doi: #1}\else
  \providecommand{\doi}{doi: \begingroup \urlstyle{rm}\Url}\fi

\bibitem[Argall et~al.(2009)Argall, Chernova, Veloso, and Browning]{argall2009survey}
Brenna~D Argall, Sonia Chernova, Manuela Veloso, and Brett Browning.
\newblock A survey of robot learning from demonstration.
\newblock \emph{Robotics and autonomous systems}, 57\penalty0 (5):\penalty0 469--483, 2009.

\bibitem[Atkeson and Schaal(1997)]{atkeson1997robot}
Christopher~G Atkeson and Stefan Schaal.
\newblock Robot learning from demonstration.
\newblock In \emph{ICML}, volume~97, pages 12--20, 1997.

\bibitem[Avigal et~al.(2022)Avigal, Berscheid, Asfour, Kr{\"o}ger, and Goldberg]{avigal2022speedfolding}
Yahav Avigal, Lars Berscheid, Tamim Asfour, Torsten Kr{\"o}ger, and Ken Goldberg.
\newblock Speedfolding: Learning efficient bimanual folding of garments.
\newblock In \emph{2022 IEEE/RSJ International Conference on Intelligent Robots and Systems (IROS)}, pages 1--8. IEEE, 2022.

\bibitem[Black et~al.(2024)Black, Brown, Driess, Esmail, Equi, Finn, Fusai, Groom, Hausman, Ichter, et~al.]{black2024pi_0}
Kevin Black, Noah Brown, Danny Driess, Adnan Esmail, Michael Equi, Chelsea Finn, Niccolo Fusai, Lachy Groom, Karol Hausman, Brian Ichter, et~al.
\newblock pi0: A vision-language-action flow model for general robot control.
\newblock \emph{arXiv preprint arXiv:2410.24164}, 2024.

\bibitem[Cadene et~al.(2024)Cadene, Alibert, Soare, Gallouedec, Zouitine, and Wolf]{cadene2024lerobot}
Remi Cadene, Simon Alibert, Alexander Soare, Quentin Gallouedec, Adil Zouitine, and Thomas Wolf.
\newblock Lerobot: State-of-the-art machine learning for real-world robotics in pytorch.
\newblock \url{https://github.com/huggingface/lerobot}, 2024.

\bibitem[Chen et~al.(2024{\natexlab{a}})Chen, Monso, Du, Simchowitz, Tedrake, and Sitzmann]{chen2024diffusion}
Boyuan Chen, Diego~Marti Monso, Yilun Du, Max Simchowitz, Russ Tedrake, and Vincent Sitzmann.
\newblock Diffusion forcing: Next-token prediction meets full-sequence diffusion.
\newblock \emph{arXiv preprint arXiv:2407.01392}, 2024{\natexlab{a}}.

\bibitem[Chen et~al.(2024{\natexlab{b}})Chen, Lim, Kelvin, Chen, and Soh]{Chen-RSS-24}
Kaiqi Chen, Eugene Lim, Lin Kelvin, Yiyang Chen, and Harold Soh.
\newblock {Don't Start From Scratch: Behavioral Refinement via Interpolant-based Policy Diffusion}.
\newblock In \emph{Proceedings of Robotics: Science and Systems}, Delft, Netherlands, July 2024{\natexlab{b}}.
\newblock \doi{10.15607/RSS.2024.XX.122}.

\bibitem[Chi et~al.()Chi, Feng, Du, Xu, Cousineau, Burchfiel, and Song]{chiDiffusionPolicyVisuomotor2023b}
Cheng Chi, Siyuan Feng, Yilun Du, Zhenjia Xu, Eric Cousineau, Benjamin Burchfiel, and Shuran Song.
\newblock Diffusion policy: Visuomotor policy learning via action diffusion.
\newblock In \emph{Robotics: Science and Systems 2023}. Robotics: Science and Systems Foundation.
\newblock ISBN 978-0-9923747-9-2.
\newblock \doi{10.15607/RSS.2023.XIX.026}.
\newblock URL \url{http://www.roboticsproceedings.org/rss19/p026.pdf}.

\bibitem[Chi et~al.(2023)Chi, Xu, Feng, Cousineau, Du, Burchfiel, Tedrake, and Song]{chi2023diffusion}
Cheng Chi, Zhenjia Xu, Siyuan Feng, Eric Cousineau, Yilun Du, Benjamin Burchfiel, Russ Tedrake, and Shuran Song.
\newblock Diffusion policy: Visuomotor policy learning via action diffusion.
\newblock \emph{The International Journal of Robotics Research}, page 02783649241273668, 2023.

\bibitem[Chi et~al.(2024)Chi, Xu, Pan, Cousineau, Burchfiel, Feng, Tedrake, and Song]{Chi-RSS-24}
Cheng Chi, Zhenjia Xu, Chuer Pan, Eric Cousineau, Benjamin Burchfiel, Siyuan Feng, Russ Tedrake, and Shuran Song.
\newblock {Universal Manipulation Interface: In-The-Wild Robot Teaching Without In-The-Wild Robots}.
\newblock In \emph{Proceedings of Robotics: Science and Systems}, Delft, Netherlands, July 2024.
\newblock \doi{10.15607/RSS.2024.XX.045}.

\bibitem[Florence et~al.(2021)Florence, Lynch, Zeng, Ramirez, Wahid, Downs, Wong, Lee, Mordatch, and Tompson]{ibc}
Pete Florence, Corey Lynch, Andy Zeng, Oscar~A Ramirez, Ayzaan Wahid, Laura Downs, Adrian Wong, Johnny Lee, Igor Mordatch, and Jonathan Tompson.
\newblock Implicit behavioral cloning.
\newblock In \emph{5th Annual Conference on Robot Learning}, 2021.

\bibitem[Florence et~al.(2019)Florence, Manuelli, and Tedrake]{florence2019self}
Peter Florence, Lucas Manuelli, and Russ Tedrake.
\newblock Self-supervised correspondence in visuomotor policy learning.
\newblock \emph{IEEE Robotics and Automation Letters}, 5\penalty0 (2):\penalty0 492--499, 2019.

\bibitem[Gupta et~al.(2019)Gupta, Kumar, Lynch, Levine, and Hausman]{gupta2019relay}
Abhishek Gupta, Vikash Kumar, Corey Lynch, Sergey Levine, and Karol Hausman.
\newblock Relay policy learning: Solving long-horizon tasks via imitation and reinforcement learning.
\newblock \emph{arXiv preprint arXiv:1910.11956}, 2019.

\bibitem[Hu et~al.(2024)Hu, Liu, Liu, and Liu]{hu2024adaflow}
Xixi Hu, Bo~Liu, Xingchao Liu, and Qiang Liu.
\newblock Adaflow: Imitation learning with variance-adaptive flow-based policies.
\newblock \emph{arXiv preprint arXiv:2402.04292}, 2024.

\bibitem[Høeg et~al.(2024)Høeg, Du, and Egeland]{høeg2024streamingdiffusionpolicyfast}
Sigmund~H. Høeg, Yilun Du, and Olav Egeland.
\newblock Streaming diffusion policy: Fast policy synthesis with variable noise diffusion models, 2024.
\newblock URL \url{https://arxiv.org/abs/2406.04806}.

\bibitem[Kim et~al.(2022)Kim, hyeon Park, Cho, and Kim]{kim2022automating}
Jigang Kim, J~hyeon Park, Daesol Cho, and H~Jin Kim.
\newblock Automating reinforcement learning with example-based resets.
\newblock \emph{IEEE Robotics and Automation Letters}, 7\penalty0 (3):\penalty0 6606--6613, 2022.

\bibitem[Lee et~al.()Lee, Wang, Etukuru, Kim, Shafiullah, and Pinto]{leebehavior}
Seungjae Lee, Yibin Wang, Haritheja Etukuru, H~Jin Kim, Nur Muhammad~Mahi Shafiullah, and Lerrel Pinto.
\newblock Behavior generation with latent actions.
\newblock In \emph{Forty-first International Conference on Machine Learning}.

\bibitem[Li and Malik(2018)]{imle}
Ke~Li and Jitendra Malik.
\newblock Implicit maximum likelihood estimation.
\newblock \emph{arXiv preprint arXiv:1809.09087}, 2018.

\bibitem[Lipman et~al.()Lipman, Chen, Ben-Hamu, Nickel, and Le]{lipmanflow}
Yaron Lipman, Ricky~TQ Chen, Heli Ben-Hamu, Maximilian Nickel, and Matthew Le.
\newblock Flow matching for generative modeling.
\newblock In \emph{The Eleventh International Conference on Learning Representations}.

\bibitem[Liu et~al.(2024{\natexlab{a}})Liu, Wu, Li, Tan, Chen, Wang, Xu, Su, and Zhu]{liu2024rdt}
Songming Liu, Lingxuan Wu, Bangguo Li, Hengkai Tan, Huayu Chen, Zhengyi Wang, Ke~Xu, Hang Su, and Jun Zhu.
\newblock Rdt-1b: a diffusion foundation model for bimanual manipulation.
\newblock \emph{arXiv preprint arXiv:2410.07864}, 2024{\natexlab{a}}.

\bibitem[Liu et~al.()Liu, Gong, et~al.]{liuflow}
Xingchao Liu, Chengyue Gong, et~al.
\newblock Flow straight and fast: Learning to generate and transfer data with rectified flow.
\newblock In \emph{The Eleventh International Conference on Learning Representations}.

\bibitem[Liu et~al.(2024{\natexlab{b}})Liu, Hamid, Xie, Lee, Du, and Finn]{liu2024bidirectional}
Yuejiang Liu, Jubayer~Ibn Hamid, Annie Xie, Yoonho Lee, Maximilian Du, and Chelsea Finn.
\newblock Bidirectional decoding: Improving action chunking via closed-loop resampling.
\newblock \emph{arXiv preprint arXiv:2408.17355}, 2024{\natexlab{b}}.

\bibitem[Lynch et~al.(2020)Lynch, Khansari, Xiao, Kumar, Tompson, Levine, and Sermanet]{lynch2020learning}
Corey Lynch, Mohi Khansari, Ted Xiao, Vikash Kumar, Jonathan Tompson, Sergey Levine, and Pierre Sermanet.
\newblock Learning latent plans from play.
\newblock In \emph{Conference on robot learning}, pages 1113--1132. PMLR, 2020.

\bibitem[Mandlekar et~al.(2020{\natexlab{a}})Mandlekar, Ramos, Boots, Savarese, Fei-Fei, Garg, and Fox]{mandlekar2020iris}
Ajay Mandlekar, Fabio Ramos, Byron Boots, Silvio Savarese, Li~Fei-Fei, Animesh Garg, and Dieter Fox.
\newblock Iris: Implicit reinforcement without interaction at scale for learning control from offline robot manipulation data.
\newblock In \emph{2020 IEEE International Conference on Robotics and Automation (ICRA)}. IEEE, 2020{\natexlab{a}}.

\bibitem[Mandlekar et~al.(2020{\natexlab{b}})Mandlekar, Xu, Mart{\'\i}n-Mart{\'\i}n, Savarese, and Fei-Fei]{mandlekar2020learning}
Ajay Mandlekar, Danfei Xu, Roberto Mart{\'\i}n-Mart{\'\i}n, Silvio Savarese, and Li~Fei-Fei.
\newblock Learning to generalize across long-horizon tasks from human demonstrations.
\newblock \emph{arXiv preprint arXiv:2003.06085}, 2020{\natexlab{b}}.

\bibitem[Mandlekar et~al.(2021)Mandlekar, Xu, Wong, Nasiriany, Wang, Kulkarni, Fei-Fei, Savarese, Zhu, and Mart{\'\i}n-Mart{\'\i}n]{robomimic}
Ajay Mandlekar, Danfei Xu, Josiah Wong, Soroush Nasiriany, Chen Wang, Rohun Kulkarni, Li~Fei-Fei, Silvio Savarese, Yuke Zhu, and Roberto Mart{\'\i}n-Mart{\'\i}n.
\newblock What matters in learning from offline human demonstrations for robot manipulation.
\newblock In \emph{5th Annual Conference on Robot Learning}, 2021.

\bibitem[O’Neill et~al.(2024)O’Neill, Rehman, Maddukuri, Gupta, Padalkar, Lee, Pooley, Gupta, Mandlekar, Jain, et~al.]{o2024open}
Abby O’Neill, Abdul Rehman, Abhiram Maddukuri, Abhishek Gupta, Abhishek Padalkar, Abraham Lee, Acorn Pooley, Agrim Gupta, Ajay Mandlekar, Ajinkya Jain, et~al.
\newblock Open x-embodiment: Robotic learning datasets and rt-x models: Open x-embodiment collaboration 0.
\newblock In \emph{2024 IEEE International Conference on Robotics and Automation (ICRA)}, pages 6892--6903. IEEE, 2024.

\bibitem[Prasad et~al.(2024)Prasad, Lin, Wu, Zhou, and Bohg]{prasad2024consistency}
Aaditya Prasad, Kevin Lin, Jimmy Wu, Linqi Zhou, and Jeannette Bohg.
\newblock Consistency policy: Accelerated visuomotor policies via consistency distillation.
\newblock \emph{arXiv preprint arXiv:2405.07503}, 2024.

\bibitem[Rahmatizadeh et~al.(2018)Rahmatizadeh, Abolghasemi, B{\"o}l{\"o}ni, and Levine]{rahmatizadeh2018vision}
Rouhollah Rahmatizadeh, Pooya Abolghasemi, Ladislau B{\"o}l{\"o}ni, and Sergey Levine.
\newblock Vision-based multi-task manipulation for inexpensive robots using end-to-end learning from demonstration.
\newblock In \emph{2018 IEEE international conference on robotics and automation (ICRA)}, pages 3758--3765. IEEE, 2018.

\bibitem[Rana et~al.(2024)Rana, Abou-Chakra, Garg, Lee, Reid, and Suenderhauf]{rana2024affordance}
Krishan Rana, Jad Abou-Chakra, Sourav Garg, Robert Lee, Ian Reid, and Niko Suenderhauf.
\newblock Affordance-centric policy learning: Sample efficient and generalisable robot policy learning using affordance-centric task frames.
\newblock \emph{arXiv preprint arXiv:2410.12124}, 2024.

\bibitem[Ravichandar et~al.(2020)Ravichandar, Polydoros, Chernova, and Billard]{ravichandar2020recent}
Harish Ravichandar, Athanasios~S Polydoros, Sonia Chernova, and Aude Billard.
\newblock Recent advances in robot learning from demonstration.
\newblock \emph{Annual review of control, robotics, and autonomous systems}, 3:\penalty0 297--330, 2020.

\bibitem[Ren et~al.(2024)Ren, Lidard, Ankile, Simeonov, Agrawal, Majumdar, Burchfiel, Dai, and Simchowitz]{dppo}
Allen~Z Ren, Justin Lidard, Lars~L Ankile, Anthony Simeonov, Pulkit Agrawal, Anirudha Majumdar, Benjamin Burchfiel, Hongkai Dai, and Max Simchowitz.
\newblock Diffusion policy policy optimization.
\newblock \emph{arXiv preprint arXiv:2409.00588}, 2024.

\bibitem[Sohl-Dickstein et~al.(2015)Sohl-Dickstein, Weiss, Maheswaranathan, and Ganguli]{sohldickstein2015nonequilibrium}
Jascha Sohl-Dickstein, Eric Weiss, Niru Maheswaranathan, and Surya Ganguli.
\newblock Deep unsupervised learning using nonequilibrium thermodynamics.
\newblock In \emph{International Conference on Machine Learning}, 2015.

\bibitem[Song et~al.(2023)Song, Dhariwal, Chen, and Sutskever]{song2023consistency}
Yang Song, Prafulla Dhariwal, Mark Chen, and Ilya Sutskever.
\newblock Consistency models.
\newblock \emph{arXiv preprint arXiv:2303.01469}, 2023.

\bibitem[Vashist et~al.(2025)Vashist, Peng, and Li]{rsimle}
Chirag Vashist, Shichong Peng, and Ke~Li.
\newblock Rejection sampling imle: Designing priors for better few-shot image synthesis.
\newblock In \emph{European Conference on Computer Vision}, pages 441--456. Springer, 2025.

\bibitem[Wang et~al.()Wang, Hart, Surovik, Kelestemur, Huang, Zhao, Yeatman, Wang, Walters, and Platt]{wangequivariant}
Dian Wang, Stephen Hart, David Surovik, Tarik Kelestemur, Haojie Huang, Haibo Zhao, Mark Yeatman, Jiuguang Wang, Robin Walters, and Robert Platt.
\newblock Equivariant diffusion policy.
\newblock In \emph{8th Annual Conference on Robot Learning}.

\bibitem[Yang et~al.()Yang, Cao, Deng, Antonova, Song, and Bohg]{yang2024equibot}
Jingyun Yang, Ziang Cao, Congyue Deng, Rika Antonova, Shuran Song, and Jeannette Bohg.
\newblock Equibot: Sim (3)-equivariant diffusion policy for generalizable and data efficient learning.
\newblock In \emph{CoRL 2024 Workshop on Whole-body Control and Bimanual Manipulation: Applications in Humanoids and Beyond}.

\bibitem[Ze et~al.(2024)Ze, Zhang, Zhang, Hu, Wang, and Xu]{ze20243d}
Yanjie Ze, Gu~Zhang, Kangning Zhang, Chenyuan Hu, Muhan Wang, and Huazhe Xu.
\newblock 3d diffusion policy.
\newblock \emph{arXiv preprint arXiv:2403.03954}, 2024.

\bibitem[Zeng et~al.(2021)Zeng, Florence, Tompson, Welker, Chien, Attarian, Armstrong, Krasin, Duong, Sindhwani, et~al.]{zeng2021transporter}
Andy Zeng, Pete Florence, Jonathan Tompson, Stefan Welker, Jonathan Chien, Maria Attarian, Travis Armstrong, Ivan Krasin, Dan Duong, Vikas Sindhwani, et~al.
\newblock Transporter networks: Rearranging the visual world for robotic manipulation.
\newblock In \emph{Conference on Robot Learning}, pages 726--747. PMLR, 2021.

\bibitem[Zhang and Gienger(2024)]{zhang2024affordance}
Fan Zhang and Michael Gienger.
\newblock Affordance-based robot manipulation with flow matching.
\newblock \emph{arXiv preprint arXiv:2409.01083}, 2024.

\bibitem[Zhang et~al.(2018)Zhang, McCarthy, Jow, Lee, Chen, Goldberg, and Abbeel]{zhang2018deep}
Tianhao Zhang, Zoe McCarthy, Owen Jow, Dennis Lee, Xi~Chen, Ken Goldberg, and Pieter Abbeel.
\newblock Deep imitation learning for complex manipulation tasks from virtual reality teleoperation.
\newblock In \emph{2018 IEEE International Conference on Robotics and Automation (ICRA)}, pages 5628--5635. IEEE, 2018.

\bibitem[Zhao et~al.()Zhao, Kumar, Levine, and Finn]{zhaoLearningFineGrainedBimanual2023}
Tony~Z. Zhao, Vikash Kumar, Sergey Levine, and Chelsea Finn.
\newblock Learning fine-grained bimanual manipulation with low-cost hardware.
\newblock volume~19.
\newblock ISBN 978-0-9923747-9-2.
\newblock URL \url{https://www.roboticsproceedings.org/rss19/p016.html}.

\bibitem[Zhao et~al.(2024)Zhao, Tompson, Driess, Florence, Ghasemipour, Finn, and Wahid]{zhao2024aloha}
Tony~Z. Zhao, Jonathan Tompson, Danny Driess, Pete Florence, Seyed Kamyar~Seyed Ghasemipour, Chelsea Finn, and Ayzaan Wahid.
\newblock {ALOHA} unleashed: A simple recipe for robot dexterity.
\newblock In \emph{8th Annual Conference on Robot Learning}, 2024.
\newblock URL \url{https://openreview.net/forum?id=gvdXE7ikHI}.

\end{thebibliography}

\end{document}